\documentclass[10pt, conference]{IEEEtran}
\usepackage[letterpaper, left=0.7in, right=0.7in, bottom=0.7in, top=0.7in]{geometry}
\usepackage{cite}
\usepackage[font=scriptsize]{caption}
\usepackage{subcaption}
\usepackage[ruled,linesnumbered]{algorithm2e}
\usepackage{algpseudocode}
\makeatletter
\def\BState{\State\hskip-\ALG@thistlm}
\makeatother

\usepackage{mathtools}

\usepackage{graphicx}
\usepackage{amsmath}
\usepackage[hidelinks]{hyperref}
\usepackage{amssymb}
\usepackage{amsthm}
\usepackage{multirow}
\newcommand\numberthis{\addtocounter{equation}{1}\tag{\theequation}}

\newtheorem*{proof*}{\hspace{0.8cm}{\it{Proof}}}

\ifCLASSINFOpdf

\else

\fi
\begin{document}
\title{  \huge A Double Q-Learning Approach for Navigation of Aerial Vehicles with Connectivity Constraint \vspace{-.2cm}}
\author{    \IEEEauthorblockN{Behzad Khamidehi and Elvino S. Sousa}
	\IEEEauthorblockA{
		Department of Electrical and Computer Engineering, University of Toronto, Canada}
	Emails: b.khamidehi@mail.utoronto.ca and es.sousa@utoronto.ca \vspace{-.3cm}}
\maketitle
\begin{abstract}

This paper studies the trajectory optimization problem for an aerial vehicle with the mission of flying between a pair of given initial and final locations. The objective is to minimize the travel time of the aerial vehicle ensuring that the communication connectivity constraint required for the safe operation of the aerial vehicle is satisfied. We consider two different criteria for the connectivity constraint of the aerial vehicle which leads to two different scenarios. In the first scenario, we assume that the maximum continuous time duration that the aerial vehicle is out of the coverage of the ground base stations (GBSs) is limited to a given threshold. In the second scenario, however, we assume that the total time periods that the aerial vehicle is not covered by the GBSs is restricted. Based on these two constraints, we formulate two trajectory optimization problems. To solve these non-convex problems, we use an approach based on the double Q-learning method which is a model-free reinforcement learning technique and unlike the existing algorithms does not need perfect knowledge of the environment. Moreover, in contrast to the well-known Q-learning technique, our double Q-learning algorithm does not suffer from the over-estimation issue. Simulation results show that although our algorithm does not require prior information of the environment, it  works well and shows near optimal performance.

\end{abstract}

\begin{IEEEkeywords}
Double Q-learning, reinforcement Learning (RL), trajectory design, unmanned aerial vehicles (UAV), cellular-connected aerial vehicles.
\end{IEEEkeywords}
%
\IEEEpeerreviewmaketitle

\section{Introduction}
\IEEEPARstart{T}{he} demands for the unmanned aerial vehicles (UAVs) have been continuously increasing in recent years. High mobility, flexible deployment, and cost effectiveness of the UAVs made it possible to adopt them in a wide range of applications \cite{R_Zhang_Survey}. One of the emerging applications is the use of UAVs in cellular networks. Based on the role that the UAVs can play, the following two scenarios are considered to integrate the UAVs into cellular networks: 1) UAV-assisted cellular communication, where UAVs are equipped with base stations (BSs) and operate as communication platforms to enhance coverage of the terrestrial users \cite{Rzhang_TComm, Halim_frontier, R_Zhang_Survey ,Rzhang_Multi,Behzad_RL, Rozhina}, and 2) cellular-enabled UAV communication, where the aerial vehicles act as users with their own missions \cite{Rzhang_TComm}. In the second scenario, the aerial vehicles are covered by the ground base stations (GBSs) throughout their flights. To operate safely, it is of great importance for the aerial vehicles to maintain reliable communication links to the GBSs. However, the current cellular networks are basically designed to cover the terrestrial mobile users. Even the radiations of the GBSs’ antennas are directed downward to support the terrestrial users. As a result, coverage is not available in all places in the sky. One way to bring connectivity to the aerial vehicles is to design appropriate trajectories for them in a way that they remain connected during their flights.

The trajectory optimization problem for the cellular-connected aerial vehicles has been investigated for different scenarios in a number of recent studies \cite{Rzhang_TComm, Behzad_PIMRC_2, Rzhang_Offloading, RZhang_ICC,Guevenc,Pollin}. In \cite{Rzhang_TComm}, the trajectory of an aerial vehicle is optimized with the objective of minimizing the travel time of the aerial vehicle while a minimum signal to noise ratio (SNR) constraint needs to be satisfied at all time instances. In \cite{Behzad_PIMRC_2}, using the successive convex approximation (SCA) technique, a trajectory optimization algorithm has been proposed to minimize the total propulsion power consumption of a fixed-wing aerial vehicle ensuring that a cellular-connectivity constraint is satisfied. In \cite{Rzhang_Offloading}, a computation offloading and trajectory optimization problem for a single UAV scenario has been investigated and an algorithm has been proposed to minimize the flight time of the UAV ensuring that the UAV is able to accomplish certain computation tasks. 
In \cite{RZhang_ICC}, using graph theory and convex optimization, the authors proposed an algorithm to minimize the UAV mission completion time. This minimization is subject to the maximum tolerable outage duration of the UAV. A similar problem has been studied in \cite{Guevenc} where a dynamic programming approach has been developed to find a sub-optimal solution for the problem.

It is worth mentioning that to formulate a trajectory optimization problem and solve it, the aforementioned studies \cite{Rzhang_TComm, Behzad_PIMRC_2, Rzhang_Offloading, RZhang_ICC,Guevenc} assume to have perfect knowledge of the environment and the optimization parameters (including channel and propagation models, interference, location of the GBSs, etc.). However, this assumption is not practical. In particular, for the networks with a large number of users or for the networks with rapid variations, this assumption is not realistic. In these cases, even if we assume that we know all the network parameters accurately, we need a significant amount of information to exchange between the components of the network which is not desirable. As a result, it is necessary to establish methods that do not need perfect knowledge of the model. 

In this paper, we study the trajectory optimization problem for an aerial vehicle whose mission is to fly between a pair of initial and final locations. Our goal is to minimize the total travel time of the aerial vehicle ensuring that the connectivity constraint of the aerial vehicle is satisfied. We consider two different criteria for the connectivity constraint of the aerial vehicle which leads to two different problems. In the first scenario, we assume that the maximum continuous time duration that the aerial vehicle is disconnected from the GBSs is less than a given threshold. However, in the second scenario, we restrict the total time periods that the UAV is flying out of the coverage of the GBSs. These problems are non-convex and difficult to solve. To resolve this issue and tackle the problems, we propose an algorithm based on the double Q-learning method which is a model-free reinforcement learning (RL) technique, and unlike the existing work in \cite{Rzhang_TComm, Behzad_PIMRC_2, Rzhang_Offloading, RZhang_ICC,Guevenc}, does not need perfect knowledge of the environment. Moreover, by decoupling action selection from action evaluation, our double Q-learning  algorithm resolves the overestimation problem of the well-known Q-learning algorithm. 

The remainder of the paper is organized as follows. Section II presents system model . The problem formulation is also discussed in this section. Section III presents an overview of the double Q-leaning approach. In section IV, we describe the trajectory optimization problem as a double Q-learning problem and propose an algorithm to solve this learning problem. Section V presents numerical results and section VI concludes the paper.

\section{System Model and Problem Formulation}

\begin{figure}[t]
	\centering
	\includegraphics[trim={0cm 0cm 0cm 0cm}, width=2.6in,keepaspectratio]{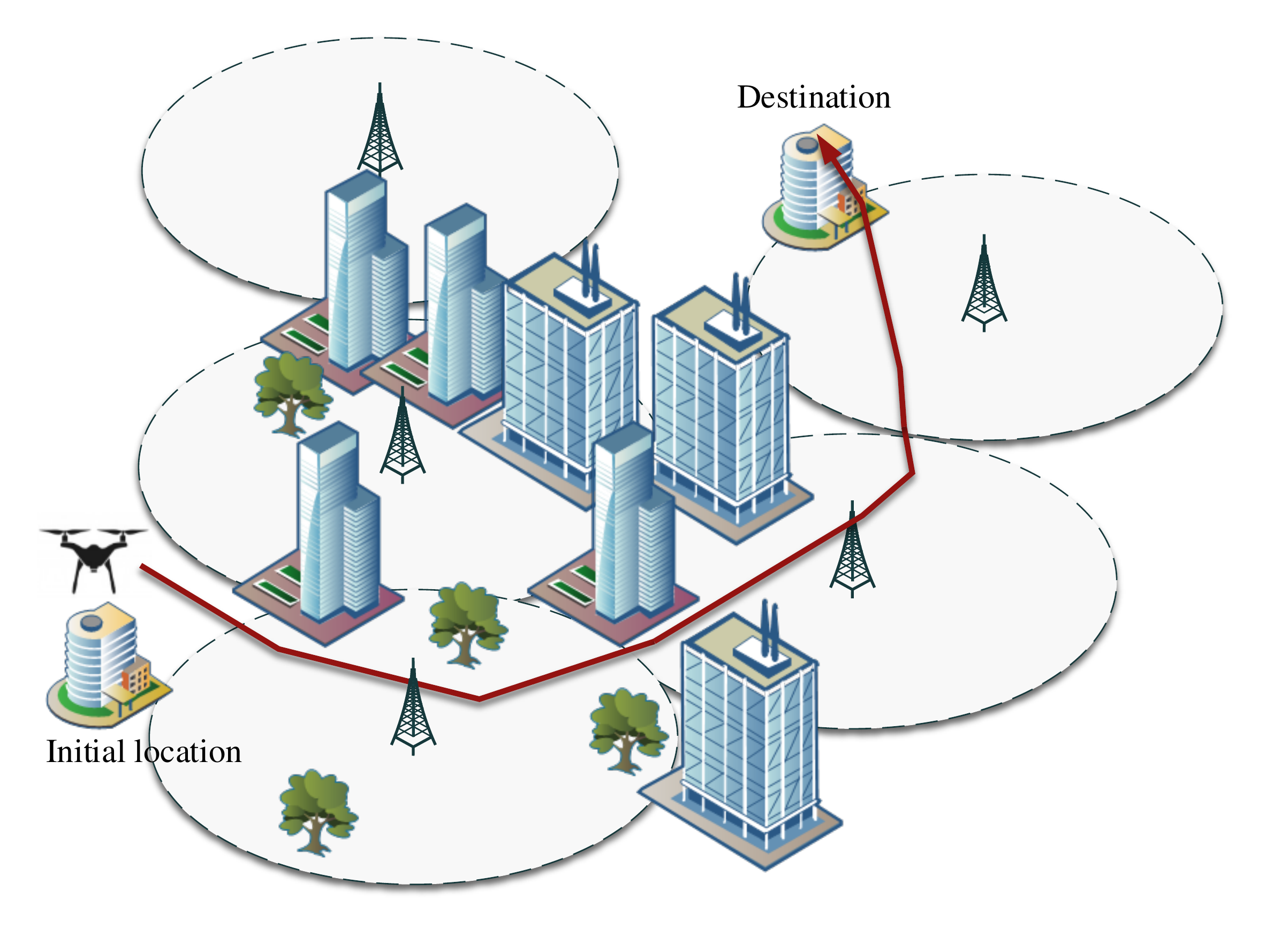}
	\vspace{-0.4em}
	\caption{A cellular-connected aerial vehicle flying between a pair of initial and final locations. }
	\label{Fig1}
	\vspace{-1.6em}
\end{figure}

As shown in Fig. \eqref{Fig1}, we consider an aerial vehicle flying over an area containing $J$ GBSs. The task of the aerial vehicle is to fly between a pair of given initial and final locations. The area of interest is denoted by $C: \mathcal{X} \times \mathcal{Y} \times \mathcal{Z}$, where $ \mathcal{X} \triangleq [x_{\text{min}},x_{\text{max}}]$, $\mathcal{Y} \triangleq [y_{\text{min}},y_{\text{max}}]$, and $\mathcal{Z} \triangleq [0,h_{\text{max}}]$, respectively. Let $q(t) = (x(t),y(t),h)$ denote the 3D position of the aerial vehicle at time $t$, where $h$ is the altitude of the aerial vehicle. We assume that the altitude $h$ is constant. The initial and final locations of the aerial vehicle are denoted by $q^{\text{I}}$ and $q^{\text{F}}$, respectively. Moreover, the aerial vehicle is not allowed to fly over {\it No-fly} zones denoted by $C^{\text{No-fly}}$. It is worth mentioning that {\it No-fly} zones are interpreted as the  areas containing obstacles at the altitude of the aerial vehicle (e.g., tall buildings), or are the regions labeled by the regulatory affairs as the banned areas. If the position of the $j$-th GBS is shown by $q_j^{\text{G}}= (x_j^{\text{G}},y_j^{\text{G}},h_j^{\text{G}})$, the distance between the aerial vehicle and this GBS at time $t$ is given by $d_j (t) = \rVert q(t) - q_j^{\text{G}} \rVert$. The speed of the aerial vehicle at time $t$ is denoted by $v(t) \triangleq \dot{q}(t)$, which is limited to the maximum speed of the aerial vehicle $v_{\text{max}}$.

\noindent The channel between the aerial vehicle and the $j$-th GBS at time $t$ is given by 
$g_j (t) = \frac{\zeta_j (t)}{\sqrt{{\text{{PL}}}_j (t)}}$,
where $\text{PL}_j (t)$ and $\zeta_j (t)$ account for the average path-loss of the communication link and the small scale fading, respectively. According to \cite{Optimal_LAP}, the average path loss expression depends on both line of sight (LoS) and Non-LoS propagations. If $\theta_j (t)$ denotes the elevation angle between the aerial vehicle and the $j$-th GBS at time $t$, the probability of having a LoS link between the aerial vehicle and the $j$-th GBS is expressed as
\begin{equation*}
\label{LoS_Pr}
\mathbb{P}_{j}^t =\frac{1}{1+a{\text{ exp}}\left(-b(\theta_{j}(t)-a)\right)},
\end{equation*} 
where $a$ and $b$ are two constants \cite{Optimal_LAP}. Hence, the average path loss $\text{PL}_j (t)$ is given by
\begin{equation*}
\label{avg_PL}
{\text{PL}}_{j}(t)= \big(\frac{4\pi f_c d_{j}(t)}{c}  \big)^2 \big(\mathbb{P}_{j}^t   \eta^{\text{LoS}} + (1-\mathbb{P}_{j}^t) \eta^{\text{N-LoS}} \big),
\end{equation*}
where $f_c$ is the carrier frequency, $c$ is the speed of light, and $\eta^{\text{LoS}}$ and  $\eta^{\text{N-LoS}}$ are additional losses for the LoS and Non-LoS links, respectively. Moreover, we assume that the small scale fading terms, $\{\zeta_j (t), \forall j, t\}$, are independent and identically distributed (iid) random variables with $\mathbb{E} \{|\zeta_j (t)|^2\}=1$. 

To show that the at time $t$, the aerial vehicle is served by the $j$-th GBS, we define variable $z_j(t)$. The value of $z_j(t)$ is 1 if the aerial vehicle is served by the $j$-th GBS at time $t$. Otherwise, $z_j(t)=0$. Using this variable, the total data rate of the aerial vehicle at time $t$ is given by 
\begin{equation}
\label{Total_rate}
R(t) = \sum_{j=1}^{J} z_j (t) \log \left( 1 + \frac{|g_j (t)|^2 p_j (t)}{I_j (t) + \sigma^2}\right), 
\end{equation}
where $p_j (t)$ denotes the transmit power of the $j$-th GBS at time $t$, $I_j(t)$ accounts for the interference arising from non-associating GBSs (GBSs other than $j$), and $\sigma^2$ is the noise power. 

\subsection{Problem Formulation}
As discussed earlier, the goal of this work is to find a trajectory for the aerial vehicle such that the travel time of the aerial vehicle between the initial and final locations is minimized. This trajectory is subject to the following constraints:

\noindent {\bf{Connectivity constraint}}: To support the command and data flows, the aerial vehicle has to maintain a reliable communication link to the GBSs. To achieve this goal, we consider two different criteria for the connectivity constraints of the aerial vehicle which leads to the following scenarios:
\begin{itemize}
	\item In the first scenario, we assume that the maximum continuous time duration that the aerial vehicle can be disconnected from the GBSs is no more than $T_{1}$ time units. Let $R_{\text{min}}$ denote the minimum required data rate for the connection of the aerial vehicle to the cellular network. To formulate this connectivity constraint, we define function $t_L (t)$ for each time instant $t$, as the last time that the aerial vehicle was connected to the cellular service, i.e., 
	$t_L (t) \triangleq \max \big\{\tau \in [0,t]: R(\tau) \geq R_{\text{min}}\big\}$.
	Using this notation, the connectivity constraint of the first scenario can be written as  
	\begin{equation}
	\label{connectivity_constraint}
	\max_{t \in [0,T]} t - t_L (t) \leq T_{1},
	\end{equation}
	where $T$ is the total travel time of the aerial vehicle and $T_1$ is the given threshold. 
\item In the second scenario, we assume that the total time periods that the aerial vehicle is out of coverage of the GBSs is less than $T_{2}$. To formulate this constraint, we define function $i(t)$ as 
\begin{equation}
\label{indicator_i}
i(t) \triangleq \mathbb{I}_{\left\{ R(t) \geq R_{\text{min}}\right\}},
\end{equation}
where $\mathbb{I}_{\{ x \}}$ is the indicator function taking value of $1$ if $x$ is true, and $0$ otherwise. Therefore, the value of function $i(t)$ is 1 if the rate of the aerial vehicle at time $t$ is greater than $R_{\text{min}}$, and is $0$, otherwise. Using this function, the connectivity constraint corresponding to the second scenario is written as
\begin{equation}
\label{connectivity_constraint_2}
\int_{0}^{T} i(t) dt \leq T_{\text{2}}.
\end{equation} 
\end{itemize}

\noindent {\bf{Initial and final locations}}: The aerial vehicle starts its mission from a given initial location and finishes its flight at a given destination, i.e., $q(0)=q^{\text{I}} \text{ and } q(T)=q^{\text{F}}$.

\noindent {\bf{Speed}}: The speed of the aerial vehicle is limited to its maximum speed, i.e., $\rVert \dot{q}(t) \rVert \leq v_{\text{max}}, \forall t \in [0,T]$.

\noindent {\bf{GBS assignment}}: At each time $t$, the aerial vehicle is connected to at most one GBS, i.e., $\sum_{j=1}^{J} z_j(t) \leq 1, \forall t \in [0,T]$.

\noindent {\bf{No-Fly zone}}: The aerial vehicle is not permitted to fly over the {\it No-fly} areas, i.e., $q(t) \notin C^{\text{No-Fly}}, \forall t \in [0,T]$. 
 
Based on the aforementioned constraints, we can formulate two optimization problems corresponding to the connectivity constraints of \eqref{connectivity_constraint} and \eqref{connectivity_constraint_2}.  If ${\bf{q}}= \{ q(t), \forall t \in [0,T]\}$, ${\bf{z}}=\{z_j (t), \forall t\in [0,T], \forall j\}$, and  $A \setminus B$ denotes the set whose elements are in $A$ but not in $B$, the trajectory optimization problem corresponding to the connectivity constraint of \eqref{connectivity_constraint} which is called {\it{scenario I}} throughout the paper will be expressed as
\begin{align*}
\label{Problem_original}
\min_{{\bf{q}}, T,{\bf{z}}}  & \hspace{0.5cm} T  \numberthis \\ 
\text{s.t. }  & \text{C1: } \max_{t \in [0,T]} t - t_L (t) \leq T_{1}\\
& \text{C2: } \begin{matrix}  \rVert \dot{q}(t) \rVert \leq v_{\text{max}}, \hspace{0.5cm} \forall t \in [0,T], \end{matrix} \\
& \text{C3: } \begin{matrix}  q(0)=q^{\text{I}}, \end{matrix} \\
& \text{C4: } \begin{matrix}  q(T)=q^{\text{F}}, \end{matrix} \\
& \text{C5: } \begin{matrix} q(t) \in C \setminus C^{\text{No-Fly}}, \hspace{0.5cm} \forall t \in [0,T], \end{matrix}\\
& \text{C6:} \sum_{j=1}^{J} z_j(t) \leq 1, \hspace{0.5cm} \forall t \in [0,T], \\
& \text{C7: } \begin{matrix}  z_j(t) \in \{0,1\}, \hspace{0.5cm} \forall t \in [0,T] \end{matrix}.
\end{align*}
The optimization problem corresponding to the connectivity constraint \eqref{connectivity_constraint_2} which is called {\it{scenario II}} is given by 
\begin{align*}
\label{Problem_original_2}
\min_{{\bf{q}}, T,z}  & \hspace{0.5cm} T  \numberthis \\ 
\text{s.t. }  & \tilde{\text{C}}\text{1: } \int_{0}^{T} i(t) dt \leq T_{2}, \hspace{0.5cm} \text{C2-C7}.
\end{align*}
\noindent The optimization problems of \eqref{Problem_original} and \eqref{Problem_original_2} are non-convex problems. To resolve this issue and tackle the problems, we have to make them convex or alternatively, approximate them in a way that the resulting problems become convex. However, due to the fact that the travel time $T$ is a variable of these optimization problems, we can not apply the successive convex approximation method utilized in our previous work \cite{Behzad_PIMRC_2}. Even if we assume there exist efficient approximations for these problems, to solve the corresponding optimization problems we need to have prior and perfect knowledge of the environment (channel gains, propagation model, interference from non-associated GBSs, location of the GBSs, etc.) which is not a valid assumption in general. Instead, we use a technique called double Q-learning which is a model-free reinforcement learning technique and does not need prior knowledge of the environment. In what follows, we briefly describe an overview of the double Q-learning method and then, we show that we can apply this approach to problems \eqref{Problem_original} and \eqref{Problem_original_2}.

\section{Double Q-learning Fundamentals}
In reinforcement learning, the agent iteratively interacts with the environment. This environment is usually described by a Markov decision process (MDP) $\prec \hspace{-0.1cm}\mathcal{S}, \mathcal{A} , \mathcal{P}, \mathcal{R}, \gamma \hspace{-0.1cm} \succ$, with space state $\mathcal{S}$, action space $\mathcal{A}$, state transition probability $\mathcal{P}(s'|s,a)$, reward function $\mathcal{R}(s,a,s')$, and discount factor $\gamma \in [0,1)$. According to this notation, at each time $t$, the agent takes an action $a_t \in \mathcal{A}$, and goes from state $s_t$ to a new state $s_{t+1}$ and receives reward of $r_{t+1}$ from the environment. If we define policy $\pi (s,a)$ as the probability of taking action $a_t=a$ in state $s_t = s$, i.e, $\pi (s,a) = {\bf{\text{Pr}}} (a_t =a | s_t =s)$, the goal of the agent is to learn a policy that maximizes the expected sum of discounted rewards it receives over the long run. This sum of discounted rewards is called return and is defined as $G_t \triangleq \sum_{k=0}^{T-1} \gamma^k r_{t+k+1}$, where $T$ is the final time step.  To evaluate and modify the implemented policy, a function is used to represent the expected return the agent receives by following the current policy for each state-action pair. This function is called state-action value function or simply Q-function and is defined as
\begin{equation}
\label{Q_func}
Q_{\pi} (s,a) \triangleq \mathbb{E} \left\{ G_t | s_t =s , a_t =a \right\},
\end{equation}
which represents the value of taking action $a$ in state $s$ under policy $\pi$. If $Q^* (s,a) = \max_{\pi} Q_{\pi} (s,a)$ denotes the optimal Q-function, the optimal policy is determined as 
\begin{equation}
\label{Optimal_Policy}
\pi^* (a|s) = \begin{cases}
1 \quad {\text{if }} a= \arg \max_a Q^* (s,a),\\
0 \quad {\text{Otherwise}}.
\end{cases}
\end{equation}
To find the optimal Q-function $Q^* (s,a)$, we have to solve the Bellman optimality equation  \cite{Sutton} as
\begin{equation*}
\label{Bellman}
Q^* (s,a) \hspace{-0.07cm}= \hspace{-0.07cm} \mathbb{E} \left\{ \hspace{-0.07cm} r_{t+1} \hspace{-0.07cm}+ \hspace{-0.07cm} \gamma \max_{a'} Q^* (s_{t+1},a')  | s_t \hspace{-0.07cm}=\hspace{-0.07cm}s , a_t \hspace{-0.07cm}=a\hspace{-0.07cm} \right\}.
\end{equation*}  
However, this equation in general is non-linear and does not have any closed form solution. Q-learning is an iterative method proposed to find $Q^* (s,a)$ \cite{Watkins}. According to the Q-learning method, upon visiting the state-action pair of $(s_t,a_t)$, the corresponding state-action value function $ Q(s_t,a_t)$ is updated as  
\begin{equation}
\label{Q_learning_Update}
Q (s_t,a_t) \hspace{-0.1cm} \leftarrow \hspace{-0.05cm} Q (s_t,a_t) + \alpha \big( r_{t+1} + \gamma \max_{a} \hspace{-0.05cm} Q(s_{t+1},a) 
 - Q(s_t,a_t) \hspace{-0.05cm} \big),
\end{equation}
where $\alpha \in (0,1]$ is the learning rate. In \eqref{Q_learning_Update}, the term $r_{t+1} + \gamma \max_{a} Q(s_{t+1},a)$ is called the temporal difference (TD) target. For evaluating this term, Q-learning uses the maximum state-action value as an approximation for the maximum expected state-action value \cite{NIPS_Double_Qlearning}. Moreover, if we reformulate the TD target as $r_{t+1} + \gamma Q(s_{t+1},\arg \max_{a} Q(s_{t+1},a))$, we can see that the max operation in the TD target uses the same values both for selecting an action and for evaluating the action. This can lead to over-estimation problem for the Q-function \cite{Deep_Double_Q}. As a result, it is highly possible that the solution of the Q-learning algorithm does not converge to the Bellman optimal solution. To prevent this issue, we can decouple action selection from action evaluation by introducing two Q-functions $Q^A (s,a)$ and $Q^B (s,a)$. Then, we update each one of these Q-functions using the value of the other one as
\begin{align*}
\label{Q_learning_Update_A}
Q^A & (s_t,a_t) \leftarrow Q^A (s_t,a_t) + \alpha \big( r_{t+1} + \\ &\gamma Q^B(s_{t+1},\arg \max_{a} Q^A(s_{t+1},a))  - Q^A(s_t,a_t) \big), \numberthis\\
\label{Q_learning_Update_B}
Q^B & (s_t,a_t) \leftarrow Q^B (s_t,a_t) + \alpha \big( r_{t+1} + \\ &\gamma Q^A(s_{t+1},\arg \max_{a} Q^B(s_{t+1},a)) 
- Q^B(s_t,a_t) \big). \numberthis
\end{align*}
In other words, the main difference between Q-learning and double Q-learning is to use two Q functions instead of one function. Moreover, at each iteration, we update only one of these two functions based on \eqref{Q_learning_Update_A} or \eqref{Q_learning_Update_B}.

\subsection{Q-Function Approximation}
\begin{figure}
	\centering
	\includegraphics[trim={0cm 0cm 0cm 0cm}, width=2.3in,keepaspectratio]{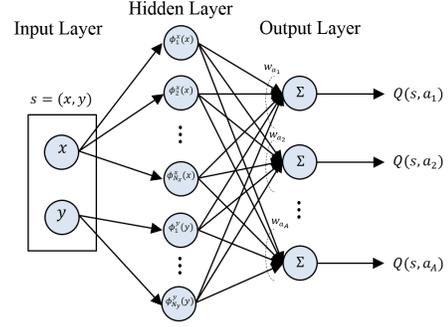}
	\vspace{-0.3em}
	\caption{Structure of the network used for the Q function approximation. }
	\label{structure}
	\vspace{-1.7em}
\end{figure}

The traditional tabular Q-learning (double Q-learning) method needs to store and update the values of the Q-function for every possible state-action pair $(s,a)$. Moreover, $Q(s,a)$ is updated only if the corresponding state-action pair $(s,a)$ is actually visited. Therefore, the tabular approach is applicable to the discrete domain problems with small number of states and actions. For problems with continuous state space or for problems with a large set of discrete states, it is not possible to use the tabular method. In these cases, we have to use function approximation to represent the Q-function as $Q(s,a;{\bf{w}}) \approx Q(s,a)$, where ${\bf{w}}$ is the parameter of our approximation. Using this representation, unlike the tabular method, we can predict the value of the Q-function for the action-state pairs which have never been visited. In this work, we use a linear function approximation for the Q-function as
\begin{equation}
\label{Q_func_approx}
Q(s,a;{\bf{w}}) = \phi (s) ^T {\bf{w}}_a,
\end{equation}
where $\phi (s)$ is the feature function and ${\bf{w}}_a$ is the weight associated with action $a$. The structure of the corresponding network used for the Q-function approximation is depicted in Fig. \eqref{structure}.

For constructing the feature function $\phi (s)$, we consider two different techniques called fixed sparse representation (FSR) and Gaussian radial basis function (RBF) as follows.
\subsubsection{Finite Sparse Representation (FSR)}
In this technique, we divide the flying area into finite regions. In particular, we divide $\mathcal{X}$ and $\mathcal{Y}$ into $N_x$ and $N_y$ spots, respectively. If we denote 
$x_k = x_{\text{min}} + \frac{k-1}{N_x} (x_{\text{max}} - x_{\text{min}})$, $\forall k =1, \ldots, N_x +1$,
and 
$y_k = y_{\text{min}} + \frac{k-1}{N_y} ( y_{\text{max}} - y_{\text{min}})$, $\forall k =1, \ldots, N_y +1$,
the feature vector $\phi (s)$ can be written as
\begin{equation}
\label{Feature_vector}
\phi (s) = \left[ \phi_{1}^x (s), \ldots, \phi_{N_x}^{x} (s), \phi_{1}^y (s), \ldots, \phi_{N_y}^{y} (s)  \right],
\end{equation}
where $\phi_k^x (s) = \begin{cases}
1 \quad \text{if } x_k \leq x < x_{k+1}, \\
0 \quad {\text{Otherwise}}, 
\end{cases}$, $\forall k = 1, \ldots, N_x$, and 
\begin{equation*}
\label{FSR_y}
\begin{matrix}
\phi_k^y (s) = \begin{cases}
1 \quad \text{if } y_k \leq y < y_{k+1}, \\
0 \quad {\text{Otherwise}}, 
\end{cases} &  \forall k = 1, \ldots, N_y.
\end{matrix}
\end{equation*} 
As a result, per action we have a parameter ${\bf{w}}_a$ which has dimension of $N_x + N_y$, and in total, we have to learn the value of $(N_x + N_y ) \times A$ parameters, where $A$ is the cardinality of the  action space, i.e., $A = |\mathcal{A}|$.

\subsubsection{Gaussian Radial Basis Functions (RBFs)}
Assume that the state of our RL framework is represented by a two-dimensional vector $s(t) = (x(t), y(t))$. For each one of $x(t)$ and $y(t)$, we consider $N_x$ and $N_y$ RBF kernels, respectively. The feature vector is again expressed as \eqref{Feature_vector}. However, we define $\phi^x_{k} (s)$ and $\phi^y_{k} (s)$ as 
\begin{align*}
\phi^x_{k} (s)= e^{- \frac{|  x - x_k | ^2}{2\mu_k ^2}},  & \forall k =1, \ldots, N_x, \\
\phi^y_{k} (s)= e^{- \frac{|  y - y_k | ^2}{2\mu_k ^2} }, &\forall k =1, \ldots, N_y, 
\end{align*}
where $x_k$ and $y_k$ are the centers of the corresponding kernels and $\mu_k^2$ is the variance of the $k$-th Gaussian kernel.

\noindent To update the parameter ${\bf{w}}$ of our approximation, we define an objective function as
\begin{equation*}
J ({\bf{w}}) = \mathbb{E}_{\pi} \left\{ \left( Q_{\pi} (s,a) - Q_{\pi} (s,a; {\bf{w}}) \right)^2   \right\},
\end{equation*}
which represents the mean-squared error between the true Q-function and the approximated one. By minimizing $J(.)$ using the stochastic gradient descent (SGD) method, ${\bf{w}}$ is updated as ${\bf{w}} = {\bf{w}} + \Delta {\bf{w}}$, where 
\begin{multline}
\Delta {\bf{w}} = - \frac{1}{2} \alpha \nabla_{\bf{w}} J({\bf{w}}) = \alpha \mathbb{E}_{\pi} \big\{ r + \gamma \max_{a'} Q_{\pi} (s',a';{\bf{w}})  \\
- Q_{\pi} (s,a; {\bf{w}})  \big\} \nabla_{\bf{w}} Q(s,a;{\bf{w}}).
\end{multline} 

\begin{algorithm}[t]
	\footnotesize
	\label{DQ_Learning}
	\caption{Connectivity constrained trajectory design based on Double Q-learning}
	Initialize ${\bf{w}}^{A}$ and ${\bf{w}}^{B}$ arbitrarily \\
	\For{$episode=1$ to Max episode}{
		According to the initial location of the UAV, initialize $s_0=q^{I}$.\\
		\For {each step of episode (time $t$)} {
			\eIf{$rand(.) < \epsilon$}{Randomely choose an action (Exploration) \\}{
				${\bf{w}}' \leftarrow \frac{1}{2}\left({\bf{w}}^A + {\bf{w}}^B\right)$\\
				Choose action $a_t=\arg \max_{a} Q(s_t,a;{\bf{w}}')$ (exploitation)\\
			}
			Take action $a_t$\\ 
			Receive the immediate reward, $r_{t+1}$ according to \eqref{reward}\\ 
			Observe the new state $s_{t+1}$\\
			Randomely update either ${\bf{w}}^{A}$ or ${\bf{w}}^{B}$\\
			\eIf{update ${\bf{w}}^{A}$ }{
				Define $a^*=\arg \max_{a} Q(s_{t+1},a;{\bf{w}}^A)$\\
				$\Delta {\bf{w}} = \alpha \Big[r_{t+1} + \gamma Q(s_{t+1},a^*;{\bf{w}}^B) -Q(s_t,a_t;{\bf{w}}^A)\Big] \nabla_{{\bf{w}}^A} Q(s_t,a_t;{\bf{w}}^A) $\\
				${\bf{w}}^A \leftarrow {\bf{w}}^A + \Delta {\bf{w}}$ \\}
			{Define $a^*=\arg \max_{a} Q(s_{t+1},a;{\bf{w}}^B)$\\
				$\Delta {\bf{w}} = \alpha \Big[r_{t+1} + \gamma Q(s_{t+1},a^*;{\bf{w}}^A) -Q(s_t,a_t;{\bf{w}}^B)\Big] \nabla_{{\bf{w}}^B} Q(s_t,a_t;{\bf{w}}^B) $\\
				${\bf{w}}^B \leftarrow {\bf{w}}^B + \Delta {\bf{w}}$ \\}
		}
	}
\end{algorithm}

\vspace{-0.2cm}
\section{Trajectory design as a double Q-learning problem}
In this section, we discuss how to use double Q-learning to problems \eqref{Problem_original} and \eqref{Problem_original_2}. As discussed earlier, in RL, the agent interacts with the environment at each of a sequence of discrete time steps $t_n, \forall n=1,2, \ldots$. In other words, instead of dealing with the continuous problem, it looks at it every $\delta t$ time units, where $\delta t$ is the time interval between two consequent time steps, i.e., $\delta t = t_n - t_{n-1}$. In what follows, we explain how to choose the value of $\delta t$ for each one of problems \eqref{Problem_original} and \eqref{Problem_original_2}: 

\begin{itemize}
	\item For problem \eqref{Problem_original}, since the UAV is not permitted to loose its connection for more than $T_{1}$ time units, it is sufficient to consider $\delta t = T_1$ and look at this problem every $T_1$ time units. If at these specific time instances, the data rate of the aerial vehicle is higher than $R_{\text{min}}$, we can guarantee that the problem is feasible and all the constraints are satisfied. 
	\item For problem \eqref{Problem_original_2}, we choose the value of $\delta t$ in a way that the environment does not change in that duration or its changes are negligible. 
\end{itemize}
Now, we can define the components of our RL framework:

\noindent {\bf{Agent}}: The agent of our problem is the aerial vehicle.\\
{\bf{State}}: The position of the aerial vehicle at each time slot is the state of the Q-learning problem, i.e., $s(t_n) = q(t_n)$.\\
{\bf{Action}}: The action is the direction of the movement, i.e., $ a(t_n ) = \Phi$. Let $D= \rVert v(t_n) \rVert \delta t (\cos \Phi {\hat{i}}+ \sin \Phi {\hat{j}})$ denote the displacement vector at each time slot, where ${\hat{i}}$ and ${\hat{j}}$ are the unit normal vectors for the $x$ and $y$ axes, respectively. Since our goal is to minimize the travel time of the aerial vehicle, it can be shown that the aerial vehicle flies with its maximum speed, i.e., $\rVert v(t) \rVert = v_{\text{max}}, \forall t \in [0,T]$. Hence, $D= v_{\text{max}} \delta t (\cos \Phi {\hat{i}}+ \sin \Phi {\hat{j}})$. Although $\Phi$ is continuous in general, in this work, we restrict its values to a limited number of angles and assume that the UAV chooses its action (direction angle) from the set of $\Phi \in \left\{0, \frac{\pi}{4}, \frac{\pi}{2}, \frac{3\pi}{4}, {\pi}, \frac{5\pi}{4}, \frac{3\pi}{2}, \frac{7\pi}{4}   \right\}$.\\
{\bf{Reward}}: The reward function is defined as follows
\begin{equation}
\label{reward}
r_{t_n} = -1 + \lambda c_{t_n} + p_{t_n},
\end{equation}
where $\lambda > 1$ is a design parameter and $p_{t_n}$ is the penalty function considered to penalize the agent if it takes action towards a point out of $C \setminus C^{\text{No-fly}}$. For problem \eqref{Problem_original}, we consider $c_{t_n}$ as 
\begin{equation}
\label{c_t}
c_{t_n} = \begin{cases}
-1 & \text{if } \quad R({t_n}) < R_{\text{min}}\\
0 & \text{if } \quad R({t_n}) \geq R_{\text{min}}.
\end{cases}
\end{equation}
However, for the problem of \eqref{Problem_original_2}, we define $c_{t_n}$ as
\begin{equation}
\label{c_t_2}
c_{t_n} = \begin{cases}
\frac{-i(t_n)}{\lambda  } & \text{if } \quad  \sum_{k=1}^{n}  i(t_k) \delta t < T_2,\\
-1 & \text{if } \quad \sum_{k=1}^{n}  i(t_k) \delta t \geq  T_2,
\end{cases}
\end{equation}
where $i(t)$ is defined in \eqref{indicator_i}.

\noindent In what follows, we describe the rationale behind the reward function of \eqref{reward}. Since our goal is to minimize the travel time of the aerial vehicle, we consider the term $-1$ as the reward to penalize the agent if it takes extra steps. Using this reward, we motivate the agent to finish its task as soon as possible. The term $c_{t_n}$ is added to penalize the agent if the connectivity constraint of the aerial vehicle is not satisfied. The third term is considered to prevent the agent from taking actions in non-feasible spaces. 

\noindent It is worth mentioning that to find $R(t_n)$ we need to know the value of $z_j (t_n)$ for $\forall j $. It can be shown that in the optimal solution of \eqref{Problem_original} and \eqref{Problem_original_2}, the optimal value of $z_j (t)$ is given by
\begin{equation}
\label{optimal_z}
z_{j}(t) = \begin{cases}
1 \quad {\text{if }} j = \arg \max_{j'} \frac{|g_{j'} (t)|^2 p_{j'} (t)}{I_{j'} (t) + \sigma^2},\\
0 \quad {\text{Otherwise}}.
\end{cases}
\end{equation}

\noindent Algorithm 1 represents our connectivity constrained trajectory design algorithm based on the double Q-leaning technique.

\section{Simulation Results}
In this section, simulation results are presented to evaluate the performance of the proposed double Q-learning trajectory design algorithm. We consider two different maps with different number of GBSs. 
The number of GBSs in map I  is $8$, and map II has $11$ GBSs. The mission of the aerial vehicle in map I is to fly from $q^{\text{I}}=(150,100)$ to $q^{\text{F}}=(900,300)$. While, in the second map the initial and final locations are $q^{\text{I}}=(100,700)$ and $q^{\text{F}}=(300,400)$, respectively. We assume that all GBSs transmit with the same power, i.e., $p_j (t)=p=200$mW, $\forall j,t $. The altitude of the aerial vehicle is fixed at $h=100$m. The carrier frequency is $2$GHz. The environment-related parameters are as follows: $a=5$, $b=0.5$, $\eta^{\text{LoS}}=1$ and $\eta^{\text{N-LoS}}=20$. The maximum velocity of the aerial vehicle is $v_{\text{max}}=10 \frac{m}{s}$. The value of $\delta t$ for the second scenario is assumed to be $0.5$s. The discount factor is $\gamma = 0.9$, $\lambda =20$, and the minimum data rate required for the connectivity of the aerial vehicle and a GBS is $R_{\text{min}}=30$bps/Hz. 

\begin{figure}[t]
	\centering
	\includegraphics[trim={0cm 0cm 0cm 0cm}, width=2.6in,keepaspectratio]{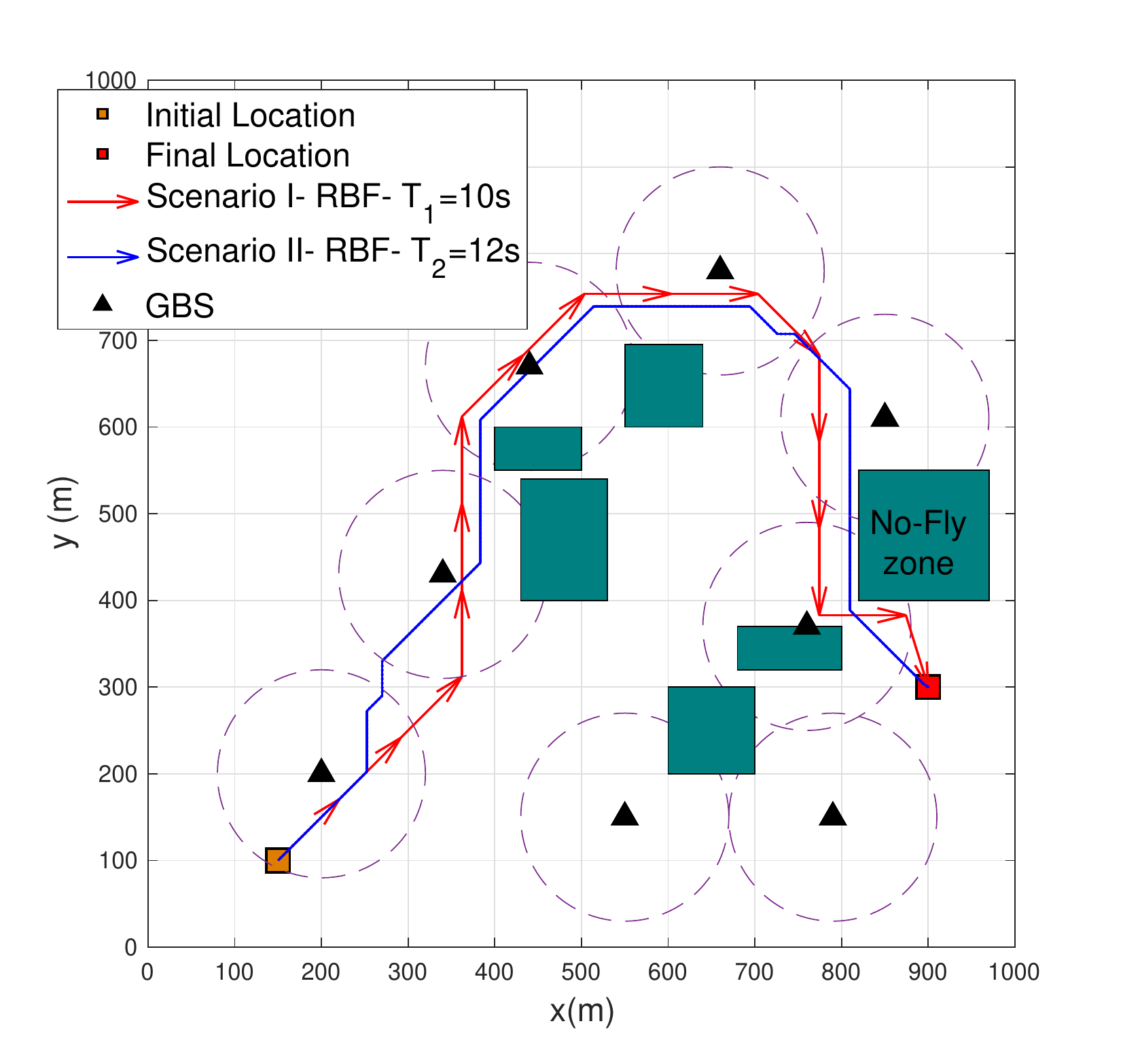}
	\vspace{-0.5em}
	\caption{ Trajectory of the UAV for map I. {\it No-Fly} areas are depicted by green rectangles.}
	\label{Traj_MAP1}
	\vspace{-1.5em}
\end{figure}%

Fig. \eqref{Traj_MAP1} shows the trajectory of the aerial vehicle for map I when RBF feature vector is used. {\it{Scenario I}} and {\it{Scenario II}} correspond to problems \eqref{Problem_original} and \eqref{Problem_original_2}, respectively. The dashed circles highlight the regions where the received data rate of the aerial vehicle is no less than $30$bps/Hz. As can be observed, the aerial vehicle starts its flight from $q^{\text{I}}$ and finishes it at $q^{\text{F}}$ while the connectivity constraint of the aerial vehicle is satisfied. It is worth mentioning that in map I, a continuous connectivity trajectory (corresponding to $T_2 =0$ in problem \eqref{Problem_original_2}) for the aerial vehicle does not exist and the minimum value of $T_2$ for having a feasible solution for \eqref{Problem_original_2} is $12$s. In fact, existence of such trajectory depends on the topology of the network and location of the GBSs, and according to the configuration of the GBSs in map I, a continuous connectivity trajectory does not exist.

\begin{figure}[t]
	\centering
	\includegraphics[trim={0cm 0cm 0cm 0cm}, width=2.6in,keepaspectratio]{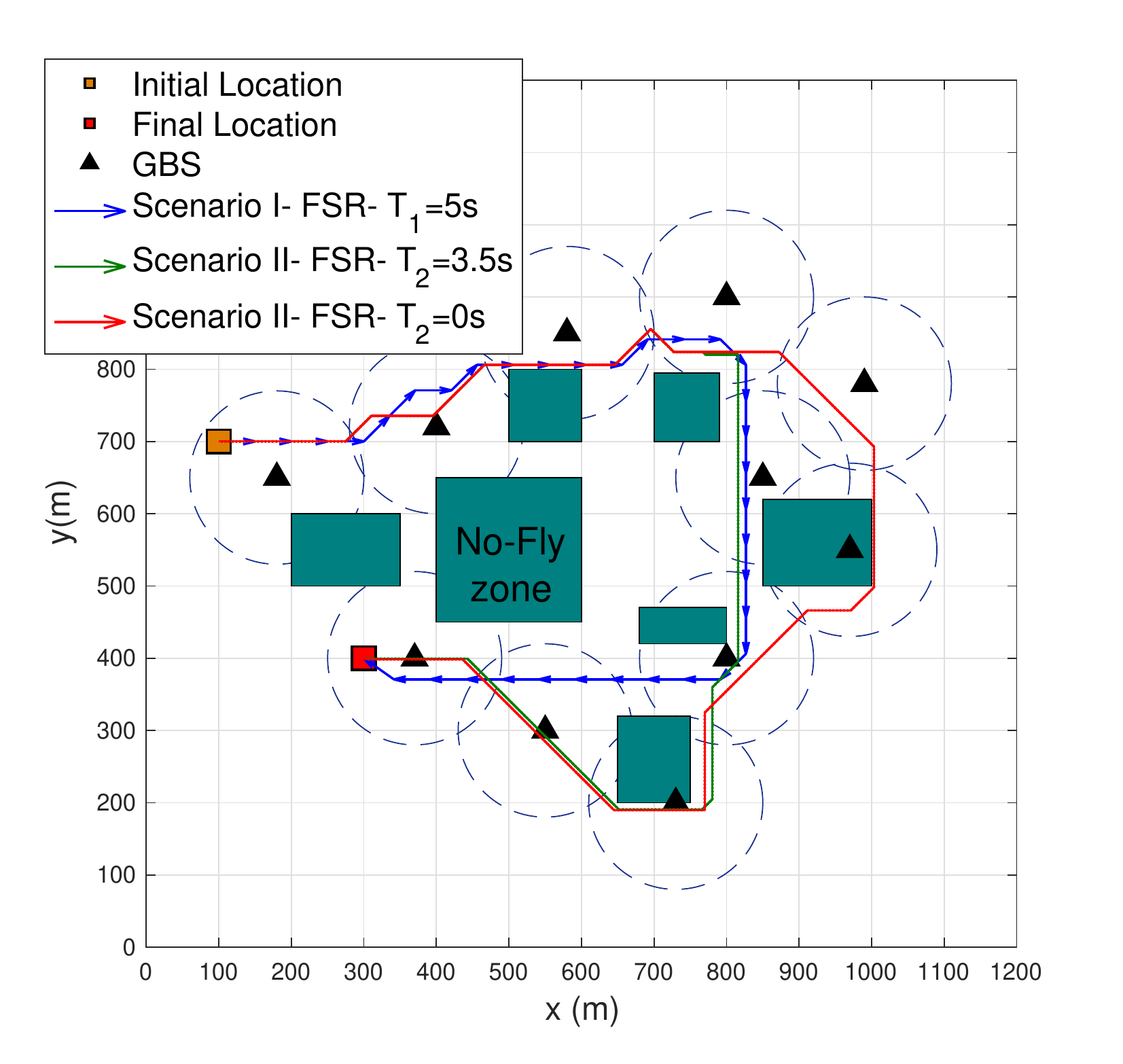}
	\vspace{-0.5em}
	\caption{Trajectory of the UAV for map II. {\it No-Fly} areas are depicted by green rectangles. }
	\label{Traj_MAP2}
	\vspace{-0.1em}
\end{figure}

The trajectory of the aerial vehicle for map II is depicted in Fig. \eqref{Traj_MAP2}. To derive these trajectories, FSR feature vector has been utilized. As can be seen, in contrast to map I, map II has a continuous connectivity trajectory ($T_2=0$). As the value of $T_2$ increases, the length of the trajectory of the aerial vehicle decreases. This is due to the fact that the aerial vehicle is allowed to fly over regions without connectivity while its connectivity constraint is not violated.

Table \ref{table_1} represents the gap between the optimal solution of \eqref{Problem_original} and \eqref{Problem_original_2} and our proposed solution when $T_1=T_2=15$s.  As can be observed, our proposed algorithm can achieve near optimal performance. This near-optimal performance is attained without having prior knowledge of the environment. As discussed earlier, our double Q-learning algorithm is a model-free learning algorithm which does not need to know the model of the environment including channel model, propagation model, {\it No-Fly} areas, etc. Instead, by getting the feedback signal from the environment, which is the reward signal, the aerial vehicle can learn the topology of the network to optimize and update its trajectory in an efficient way.

\section{Conclusion}

In this paper, we investigated the trajectory optimization problem for an aerial vehicle with connectivity constraint. We considered two different criteria for the connectivity constraint of the aerial vehicle which leads to two different optimization problems. In the first problem, the aerial vehicle is not allowed to loose its connection to the GBSs for more than $T_1$ time units. In the second problem, however, we restrict the total disconnection time of the aerial vehicle to $T_2$ time units. To solve these non-convex problems, we adopted an algorithm based on the double Q-learning technique which avoids the over-estimation problem of the well-known Q-learning technique by decoupling action selection from action evaluation. Moreover, our algorithm does not need perfect knowledge of the environment. 

\begin{table}[!t]
	\caption{The gap between our proposed solution and the optimal one when $T_1=T_2=15$s.}
	\label{table_1}
	\centering
	\scriptsize
	\begin{tabular}{c|c|c||c|c}
		& \multicolumn{2}{c||}{\it{Scenario I }} & \multicolumn{2}{c}{\it{Scenario II}} \\
		\hline
		& FSR & RBF & FSR & RBF\\
		\hline
		Map I &  $8.5\%$  & $9.5\%$  &  $7.5\%$  & $8\%$  \\
		\hline
		Map II &  $7\%$  &  $8\%$ &  $7\%$  &  $8\%$ \\
	\end{tabular}
	\vspace{-.5cm}
\end{table}

\bibliographystyle{IEEEtran}
\bibliography{Citations}

\begin{thebibliography}{10}
\providecommand{\url}[1]{#1}
\csname url@samestyle\endcsname
\providecommand{\newblock}{\relax}
\providecommand{\bibinfo}[2]{#2}
\providecommand{\BIBentrySTDinterwordspacing}{\spaceskip=0pt\relax}
\providecommand{\BIBentryALTinterwordstretchfactor}{4}
\providecommand{\BIBentryALTinterwordspacing}{\spaceskip=\fontdimen2\font plus
\BIBentryALTinterwordstretchfactor\fontdimen3\font minus
  \fontdimen4\font\relax}
\providecommand{\BIBforeignlanguage}[2]{{%
\expandafter\ifx\csname l@#1\endcsname\relax
\typeout{** WARNING: IEEEtran.bst: No hyphenation pattern has been}%
\typeout{** loaded for the language `#1'. Using the pattern for}%
\typeout{** the default language instead.}%
\else
\language=\csname l@#1\endcsname
\fi
#2}}
\providecommand{\BIBdecl}{\relax}
\BIBdecl

\bibitem{R_Zhang_Survey}
Y.~{Zeng}, R.~{Zhang}, and T.~J. {Lim}, ``{Wireless Communications with
  Unmanned Aerial Vehicles: Opportunities and Challenges},'' \emph{IEEE Commun.
  Mag.}, vol.~54, no.~5, pp. 36--42, May 2016.

\bibitem{Rzhang_TComm}
S.~{Zhang}, Y.~{Zeng}, and R.~{Zhang}, ``{Cellular-Enabled UAV Communication: A
  Connectivity-Constrained Trajectory Optimization Perspective},'' \emph{IEEE
  Trans. Commun.}, vol.~67, no.~3, pp. 2580--2604, Mar. 2019.

\bibitem{Halim_frontier}
I.~{Bor-Yaliniz} and H.~{Yanikomeroglu}, ``{The New Frontier in RAN
  Heterogeneity: Multi-Tier Drone-Cells},'' \emph{IEEE Commun. Mag.}, vol.~54,
  no.~11, pp. 48--55, Nov. 2016.

\bibitem{Rzhang_Multi}
Q.~{Wu}, Y.~{Zeng}, and R.~{Zhang}, ``{Joint Trajectory and Communication
  Design for Multi-UAV Enabled Wireless Networks},'' \emph{IEEE Trans. Wireless
  Commun.}, vol.~17, no.~3, pp. 2109--2121, Mar. 2018.

\bibitem{Behzad_RL}
B.~{Khamidehi} and E.~S. {Sousa}, ``{Reinforcement Learning-Based Trajectory
  Design for the Aerial Base Stations},'' in \emph{Proc. IEEE PIMRC}, Sep.
  2019, pp. 1--6.

\bibitem{Rozhina}
R.~{Ghanavi}, E.~{Kalantari}, M.~{Sabbaghian}, H.~{Yanikomeroglu}, and
  A.~{Yongacoglu}, ``{Efficient 3D aerial base station placement considering
  users mobility by reinforcement learning},'' in \emph{Proc. IEEE WCNC}, Apr.
  2018, pp. 1--6.

\bibitem{Behzad_PIMRC_2}
B.~{Khamidehi} and E.~S. {Sousa}, ``{Power Efficient Trajectory Optimization
  for the Cellular-Connected Aerial Vehicles},'' in \emph{Proc. IEEE PIMRC},
  Sep. 2019, pp. 1--6.

\bibitem{Rzhang_Offloading}
\BIBentryALTinterwordspacing
X.~Cao, J.~Xu, and R.~Zhang, ``{Mobile Edge Computing for Cellular-Connected
  {UAV:} Computation Offloading and Trajectory Optimization}.'' [Online].
  Available: \url{http://arxiv.org/abs/1803.03733}
\BIBentrySTDinterwordspacing

\bibitem{RZhang_ICC}
S.~{Zhang} and R.~{Zhang}, ``{Trajectory Design for Cellular-Connected UAV
  Under Outage Duration Constraint},'' in \emph{Proc. IEEE ICC}, May 2019, pp.
  1--6.

\bibitem{Guevenc}
E.~{Bulut} and I.~{Guevenc}, ``{Trajectory Optimization for Cellular-Connected
  UAVs with Disconnectivity Constraint},'' in \emph{Proc. IEEE ICC Workshops},
  May 2018, pp. 1--6.

\bibitem{Pollin}
S.~{De Bast}, E.~{Vinogradov}, and S.~{Pollin}, ``{Cellular Coverage-Aware Path
  Planning for UAVs},'' in \emph{Proc. IEEE SPAWC}, Jul. 2019, pp. 1--5.

\bibitem{Optimal_LAP}
A.~{Al-Hourani}, S.~{Kandeepan}, and S.~{Lardner}, ``{Optimal LAP Altitude for
  Maximum Coverage},'' \emph{IEEE Wireless Commun. Lett.}, vol.~3, no.~6, pp.
  569--572, Dec. 2014.

\bibitem{Sutton}
R.~S. Sutton and A.~G. Barto, \emph{Introduction to Reinforcement Learning},
  1st~ed.\hskip 1em plus 0.5em minus 0.4em\relax Cambridge, MA, USA: MIT Press,
  1998.

\bibitem{Watkins}
C.~J. C.~H. Watkins and P.~Dayan, ``{Q-Learning},'' \emph{Mach. Learn.},
  vol.~8, no.~3, pp. 279--292, May 1992.

\bibitem{NIPS_Double_Qlearning}
H.~V. Hasselt, ``{Double Q-learning},'' in \emph{Proc. NeurIPS}, 2010, pp.
  2613--2621.

\bibitem{Deep_Double_Q}
H.~V. Hasselt, A.~Guez, and D.~Silver, ``{Deep Reinforcement Learning with
  Double Q-Learning},'' in \emph{Proc. 13th AAAI Conf. Artif. Intell.}, 2016,
  pp. 2094--2100.

\end{thebibliography}
\end{document}